\documentclass{article}
\usepackage{ijcai22}

\usepackage{times}

\usepackage{soul}
\usepackage{url}
\usepackage[hidelinks]{hyperref}
\usepackage[utf8]{inputenc}
\usepackage[small]{caption}
\usepackage{graphicx}
\usepackage{amsmath}
\usepackage{booktabs}
\usepackage{algorithm}
\usepackage{algorithmic}
\usepackage{nicefrac}

\usepackage{amsmath}
\usepackage{amssymb}
\usepackage[small]{subfigure}
\usepackage{bbm}

\DeclareMathOperator*{\argmax}{argmax}   

\newcommand{\citet}[1]{\citeauthor{#1}~\shortcite{#1}}
\newcommand{\citep}{\cite}

\usepackage{blkarray, bigstrut}
\usepackage{enumitem}
\setlist[enumerate,itemize]{itemsep=0.5pt,topsep=0pt}

\usepackage[textsize=small,textwidth=3cm]{todonotes}
\definecolor{kentuckyblue}{RGB}{0, 93, 170}

\usepackage{enumitem}
\setlist[enumerate,itemize]{itemsep=0.5pt,topsep=0pt}

\title{Combining Fast and Slow Thinking\\ for\\ Human-like and Efficient Navigation in Constrained Environments}

\author{
Marianna B. Ganapini$^1$\and
Murray Campbell$^2$\and 
Francesco Fabiano$^3$\and 
Lior Horesh$^2$ \and  
Jon Lenchner$^2$ \and 
Andrea Loreggia$^4$ \and 
Nicholas Mattei$^5$ \and  
Taher Rahgooy$^6$ \and  
Francesca Rossi$^2$ \and 
Biplav Srivastava$^7$  \and 
Brent Venable$^8$
\affiliations
$^1$Union College \\
$^2$IBM Research\\
$^3$University of Udine\\
$^4$University of Brescia\\
$^5$Tulane University\\
$^6$ University of West Florida\\
$^7$Univ. of South Carolina\\
$^8$Institute for Human and Machine Cognition 
\emails
 bergamam@union.edu, 
\{mcam,lhoresh,lenchner,Francesca.Rossi2\}@us.ibm.com
francesco.fabiano@uniud.it,
 andrea.loreggia@gmail.com,
nsmattei@tulane.edu,
trahgooy@students.uwf.edu,
biplav.s@sc.edu,
bvenable@ihmc.org
}

\begin{document}

\maketitle

\begin{abstract}
Current AI systems lack several important human capabilities, such as adaptability, generalizability, self-control, consistency, common sense, and causal reasoning. We believe that existing cognitive theories of human decision making, such as the thinking fast and slow theory, can provide insights on how to advance AI systems towards some of these capabilities. In this paper, we propose a general architecture that is based on fast/slow solvers and a metacognitive component. We then present experimental results on the behavior of an instance of this architecture, for AI systems that make decisions about navigating in a constrained environment. We show how combining the fast and slow decision modalities allows the system to evolve over time and gradually pass from slow to fast thinking with enough experience, and that this greatly helps in decision quality, resource consumption, and efficiency.
\end{abstract}

\section{Introduction} 

AI systems have seen great advancement in recent years, on many applications that  pervade our everyday life. However, we are still mostly seeing instances of narrow AI that are 
typically focused on a very limited set of competencies and goals, e.g., image interpretation, natural language processing, classification, prediction, and many others. Moreover, while these successes can be accredited to improved algorithms and techniques, they are also tightly linked to the availability of huge datasets and computational power \cite{marcus2020next}. State-of-the-art AI still lacks many capabilities that would naturally be included in a notion of (human) intelligence, such as 
generalizability, adaptability, robustness, explainability, causal analysis, abstraction, common sense reasoning, ethical reasoning \cite{RoMa19a}, as well as a complex and seamless integration of learning and reasoning supported by both implicit and explicit knowledge \cite{ai1002021}.

We believe that a better study of the mechanisms that allow humans to have these capabilities can help \cite{aaai2021-blue}.
We focus especially on D. Kahneman's theory of thinking fast and slow  \cite{kahneman2011thinking}, and we propose a multi-agent AI architecture (called SOFAI, for SlOw and Fast AI) where incoming problems are solved by either System 1 (or ``fast") agents (also called ``solvers"), that react by exploiting only past experience, or by System 2 (or ``slow") agents, that are deliberately activated when there
is the need to reason and search for optimal solutions beyond what is expected from the System 1 agents. 
Given the need to choose between these two kinds of solvers, a meta-cognitive agent is employed, performing introspection and arbitration roles, and assessing the  need to employ System 2 solvers by considering resource constraints, abilities of the solvers, past experience, and expected reward for a correct solution of the given problem  \cite{Shenhav2013,Thompson2011}.
Different approaches to the design of AI systems inspired by the dual-system theory have also been published recently
\cite{bengio2017consciousness,goel2017thinking,chen2019deep,anthony2017thinking,mittal2017thinking,DBLP:journals/ibmrd/NoothigattuBMCM19,gulati2020interleaving}, showing that this theory inspires many AI researchers. 


In this paper we describe the SOFAI architecture, characterizing the System 1 and System 2 solvers and the role of the meta-cognitive agent, and provide motivations for the adopted design choices. We then focus on a specific instance of the SOFAI architecture, that provides the multi-agent platform for generating trajectories in a grid environment with penalties over states, actions, and state features. In this instance, decisions are at the level of each move from one grid cell to another. 
We show that the combination of fast and slow decision modalities allows the system to create trajectories that are similar to human-like ones, compared to using only one of the modalities. Human-likeness is here exemplified by the trajectories built by a Multi-alternative Decision Field Theory model (MDFT) \cite{roe2001multialternative}, that has been shown to mimick the way humans decide among several alternatives. In our case, the possible moves in a grid state, take into account non-rational behaviors related to alternatives' similarity. Moreover, the SOFAI trajectories are shown to generate a better reward and to require a shorter decision time overall. 
We also illustrate the evolution of the behavior of the SOFAI system over time, showing that, just like in humans, initially the system mostly uses the System 2 decision modality, and then passes to using mostly System 1 when enough experience over moves and trajectories is collected.

\section{Background}

We introduce the main ideas of the thinking fast and slow theory. We also describe the main features of the  Multi-alternative Decision Field Theory (MDFT) \cite{roe2001multialternative}, that we will use in the experiments (Section \ref{sec:sofai_def} and \ref{sec:experiments}) to generate human-like trajectories in the grid environment.

\subsection{Thinking Fast and Slow in Humans} 
\label{fshumans}

According to Kahneman's theory, described in his book ``Thinking, Fast and Slow" \cite{kahneman2011thinking},
human's decisions are supported and guided by the cooperation of two kinds of capabilities, that for the sake of simplicity are called \emph{systems}: System 1 (``thinking fast") provides tools for intuitive, imprecise, fast, and often unconscious decisions, while System 2 (``thinking slow") handles more complex situations where logical and rational thinking is needed to reach a complex decision.

System 1 is guided mainly by intuition rather than deliberation. It gives fast answers to simple questions. Such answers are sometimes wrong, mainly because of unconscious bias or because they rely on heuristics or other short cuts \cite{gigerenzer2009}, and usually do not provide explanations. 
However, System 1 is able to build models of the world that, although inaccurate and imprecise, can fill knowledge gaps through causal inference, allowing us to respond reasonably well to the many stimuli of our everyday life.

When the problem is too complex for System 1, System 2 kicks in and solves it with access to additional computational resources, full attention, and sophisticated logical reasoning. A typical example of a problem handled by System 2 is solving a complex arithmetic calculation, or a multi-criteria optimization problem.
To do this, humans need to be able to recognize that a problem goes beyond a threshold of cognitive ease and therefore see the need to activate a more global and accurate reasoning machinery \cite{kahneman2011thinking}. Hence, introspection and meta-cognition is essential in this process.

When a problem is new and difficult to solve, it is handled by System 2 \cite{Kim2019}. However, certain problems, over time as more experience is acquired, pass on to System 1. The procedures System 2 adopts to find solutions to such problems become part of the experience that System 1 can later use with little effort. Thus, over time, some problems, initially solvable only by resorting to System 2 reasoning tools, can become manageable by System 1. A typical example is reading text in our own native language. However, this does not happen with all tasks. An example of a problem that never passes to System 1 is finding the correct solution to complex arithmetic questions.

\subsection{Multi-Alternative Decision Field Theory}
\label{sec:mdft-back}

\begin{figure}[h]
\centering
\includegraphics[width=0.45\textwidth]{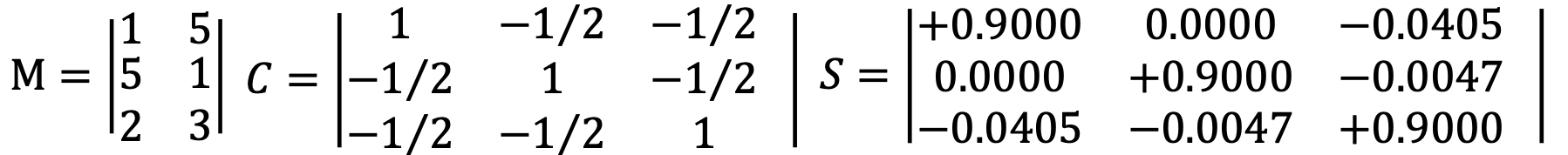}
\caption{Evaluation (M), Contrast (C), and Feedback (S) matrix.}
\label{fig:matrix}
\vspace{-1em}
\end{figure}

Multi-alternative Decision Field Theory (MDFT) 
\cite{roe2001multialternative}
models human preferential choice as an iterative cumulative process. 
In MDFT, an agent is confronted with multiple options and equipped with an initial personal evaluation for them along different criteria, called attributes. For example, a student who needs to choose a main course among those offered by the cafeteria will have in mind an initial evaluation of the options in terms of how tasty and healthy they look. More formally, MDFT comprises of:

\textbf{Personal Evaluation}: 
Given set of options $O=\{o_1, \dots, o_k\}$ and set of attributes
$A=\{A_1, \dots, A_J\}$,
the subjective value of option $o_i$ on 
attribute $A_j$ is denoted by $m_{ij}$ and stored in matrix $\textbf{M}$. 
In our example, let us assume that the cafeteria options  are 
{\em Salad (S)}, {\em Burrito (B)} and {\em Vegetable pasta (V)}.
Matrix $\mathbf{M}$, containing the student's preferences,
could be defined as shown in Figure \ref{fig:matrix} (left),
where rows correspond to the options $(S,B,V)$ and the columns 
to the attributes $Taste$ and $Health$.

\textbf{Attention Weights}: Attention weights are used to express the attention allocated to each attribute at a particular time $t$ during the deliberation. We denote them by vector $\textbf{W}(t)$ where $W_j(t)$ represents the attention to attribute $j$ at time $t$. We adopt the common simplifying assumption that, at each point in time, the decision maker attends to only one attribute \cite{roe2001multialternative}. Thus, $W_j(t)\in\{0,1\}$ and $\sum_jW_j(t)=1$, $\forall t,j$. In our example, we have two attributes, so at any point in time $t$ we will have $\textbf{W}(t)=[1,0]$, or $\textbf{W}(t)=[0,1]$, representing that the student is attending to, respectively, $Taste$ or $Health$. The attention weights change across time according to a stationary stochastic process with probability distribution $\textbf{w}$, where $w_j$ is the probability of attending to attribute $A_j$. In our example, defining $w_1=0.55$ and $w_2=0.45$ would mean that at each point in time, the student will be attending $Taste$ with probability $0.55$ and $Health$ with probability $0.45$. 

\textbf{Contrast Matrix}: Contrast matrix $\textbf{C}$ is used to 
compute the advantage (or disadvantage) of  an option with respect to the other options.
In the MDFT literature \cite{busemeyer1993decision,roe2001multialternative,hotaling2010theoretical}, $\textbf{C}$
is defined by contrasting the initial evaluation of one alternative against the 
average of the evaluations of the others, as shown for the case with three options in Figure \ref{fig:matrix} (center).


At any moment in time, each alternative in the 
choice set is associated with a {\bf valence} value. The valence for option $o_i$ 
at time $t$, denoted $v_i(t)$, represents its momentary advantage (or 
disadvantage) when compared with other options on some attribute 
under consideration. The valence vector for $k$ options $o_1, \dots, o_k$ at 
time $t$, denoted by column vector $\mathbf{V}(t) = [v_1(t), \dots, v_k(t)]^T$, is formed by $\textbf{V}(t) = \textbf{C}\times \textbf{M} \times \textbf{W}(t)$.
%
%
In our example, the valence vector at any time point in which $\textbf{W}(t)=[1,0]$, is 
$\textbf{V}(t) = [1-\nicefrac{7}{2}, 5-\nicefrac{3}{2}, 2-\nicefrac{6}{2}]^T$.

Preferences for each option are accumulated across the iterations of the deliberation process until a decision is made. This is done by using \textbf{Feedback Matrix} $\mathbf{S}$, which defines how the accumulated preferences affect the preferences computed at the next iteration. This interaction depends on how similar the options are in terms of their initial evaluation expressed in $\mathbf{M}$. Intuitively, the new preference of an option is affected positively and strongly by the preference it had accumulated so far, while it is inhibited by the preference of similar options. This 
lateral inhibition decreases as the dissimilarity between options increases. Figure \ref{fig:matrix} (right) shows 
\textbf{S} for our example \cite{hotaling2010theoretical}. 
At any moment in time, the  preference of each alternative is calculated by 
$\textbf{P}(t + 1) = \textbf{S} \times \textbf{P}(t) + \textbf{V}(t + 1)$
where  $\textbf{S} \times \textbf{P}(t)$ is the contribution of the past preferences and 
$\textbf{V}(t + 1)$ is the valence computed at that iteration. Starting with $\textbf{P}(0)=0$,
preferences are then accumulated for either a fixed number of iterations (and the option with the highest preference is selected) or until the preference of an option reaches a given threshold. 
In the first case, MDFT models decision making with a \textit{specified} deliberation time, while, in the latter, it models cases where deliberation time is \textit{unspecified} and choice is dictated by the accumulated preference magnitude.
In general, different runs of the same MDFT model 
may return different choices due to the attention weights' distribution. 
In this way, MDFT induces choice distributions over set of options and is capable of capturing well know behavioral effects such as the compromise, similarity, and attraction effects that have been observed in humans and that violate rationality principles \cite{busemeyer1993decision}. 

\section{Thinking Fast and Slow in AI} 
\label{sec:sofai}

\begin{figure}
  \centering
  \includegraphics[scale=0.26]{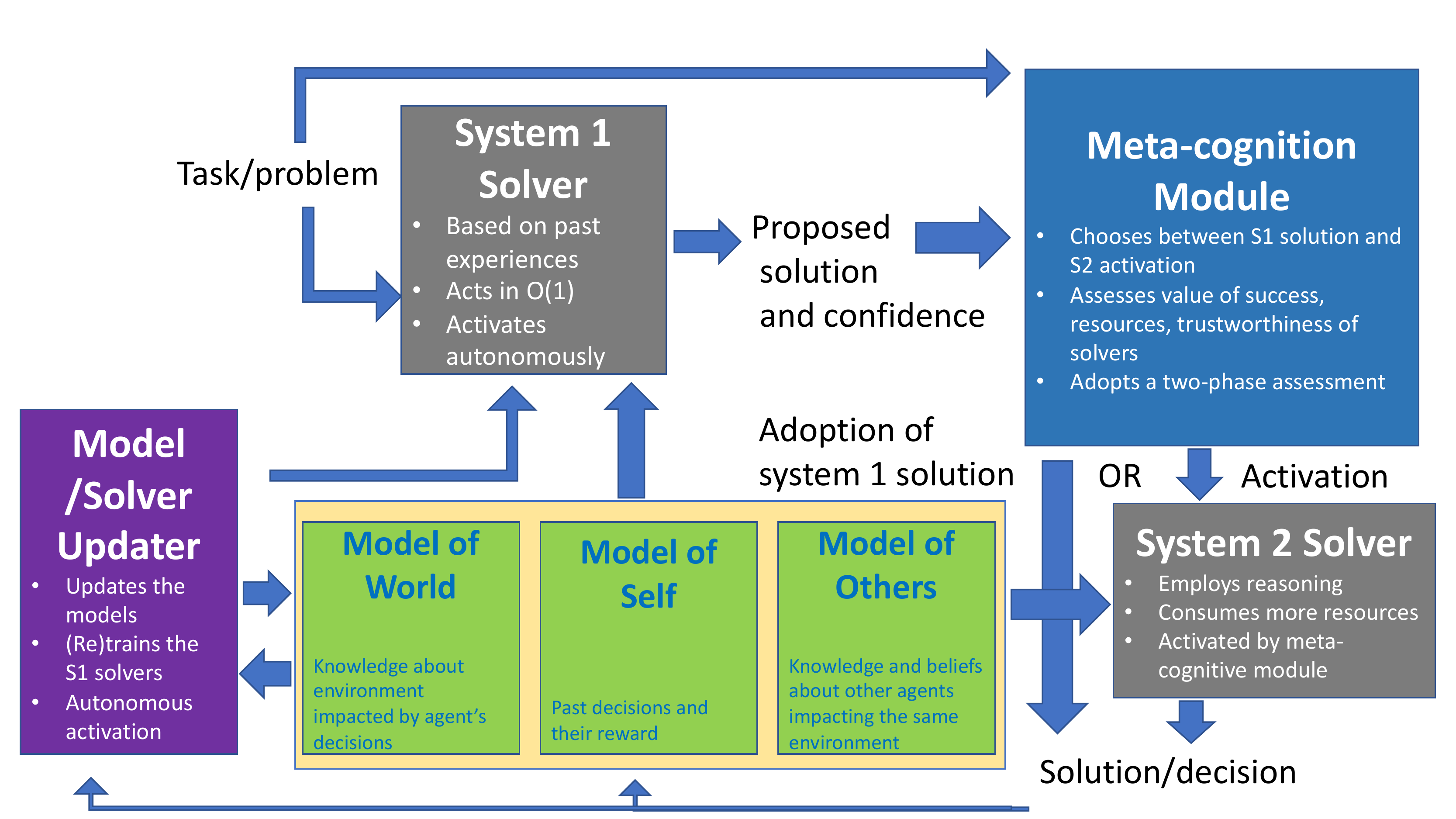}
  \caption{The SOFAI architecture.}
  \label{fig1}
\end{figure}

SOFAI is a multi-agent architecture (see Figure \ref{fig1}) where incoming problems are initially handled by those System 1 (S1) solvers that possess the required skills to tackle them, analogous to what is done by humans who first react to an external stimulus via their System 1. 

\subsection{Fast and Slow Solvers}

As mentioned, incoming problems trigger System 1 (S1) solvers. We assume such solvers act in constant time, i.e., their running time is not a function of the size of the input problem instance, by relying on the past experience of the system, which is maintained in the model of self. 
The model of the world contains the knowledge accumulated by the system over the external environment and the expected tasks, while the model of others contains the knowledge and beliefs about other agents who may act in the same environment. The model updater agent acts in the background to keep all models updated as new knowledge of the world, of other agents, or new decisions are generated and evaluated.

Once an S1 solver has solved the problem (for the sake of simplicity, assume a single S1 solver), the proposed solution and the associated confidence level are available to the meta-cognitive (MC) module. At this point the MC agent starts its operations, with the task of choosing between adopting the S1 solver's solution or activating a System 2 (S2) solver. 
S2 agents use some form of reasoning over the current problem and usually consume more resources (especially time) than S1 agents. Also, they never work on a problem unless they are explicitly invoked by the MC module. 

To make its decision, the MC agent assesses the current resource availability, the expected resource consumption of the S2 solver, the expected reward for a correct solution for each of the available solvers, as well as the solution and confidence evaluations coming from the S1 solver.
In order to not waste resources at the meta-cognitive level, the MC agent includes two successive assessment phases, the first one faster and more approximate, related to rapid unconscious assessment in humans \cite{Ackerman2017,Proust2004}, and the second one (to be used only if needed) more careful and resource-costly, analogous to the conscious introspective process in humans \cite{Carruthers}. 
The next section will provide more details about the internal steps of the MC agent. 




This architecture and flow of tasks allows for minimizing time to action when there is no need for S2 processing since S1 solvers act in constant time. It also
allows the MC agent to exploit the proposed action and confidence of S1
when deciding whether to activate S2, which leads to more informed and hopefully better decisions by the MC.

Notice that 
we do not assume that S2 solvers are always better than S1 solvers,
analogously to what happens in human reasoning \cite{gigerenzer2009}.
Take for example complex arithmetic, which usually requires humans to employ System 2, vs 
perception tasks, which are typically handled by our System 1. Similarly, in the SOFAI architecture we allow for tasks that might be better handled by S1 solvers, especially once the system has acquired enough experience on those tasks.

\subsection{The Role of Meta-cognition} 

We focus on the concept of meta-cognition as initially defined by \citet{Flavell1979,NN1990}, that is, the set of processes and mechanisms that could allow a computational system to both monitor and control its own cognitive activities, processes, and structures. The goal of this form of control is to improve the quality of the system’s decisions \cite{CoxRaja2011}. 
Among the existing computational models of meta-cognition \cite{Cox2005,Kralik2018,Posner2020},
we propose a centralized meta-cognitive module that exploits both internal and external data, and arbitrates between S1 and S2 solvers in the process of solving a single task. Notice however that this arbitration is different from an algorithm portfolio selection, 
which is already successfully used to tackle many problems \cite{kerschke2019automated},
because of the characterization of S1 and S2 solvers and the way the MC agent controls them.

The MC module 
exploits information coming from two main sources: 
1) the system’s internal models of self, world, and others;
2) the S1 solver(s), providing a proposed decision 
    for a task, and  
    their confidence in the proposed decision. 


The first meta-cognitive phase (MC1) 
activates automatically as a new task arrives and a solution for the problem is provided by an S1 solver. 
MC1 decides between accepting the solution proposed by the S1 solver or activating the second meta-cognitive phase (MC2).
MC2 then makes sure that there are enough resources for running S2. If not, MC2 adopts the S1 solver's proposed solution.
MC1 also compares the confidence provided by the S1 solver with the risk attitude of the system: if the confidence is high enough, MC1 adopts the S1 solver’s solution. Otherwise, it activates the next assessment phase (MC2) to make a more careful decision.
The rationale for this phase of the decision process is that we envision that often the system will adopt the solution proposed by the S1 solver, because it is good enough given the expected reward for solving the task, or because there are not enough resources to invoke more complex reasoning. 

Contrarily to MC1, MC2 decides between accepting the solution proposed by the S1 solver or activating an S2 solver for the task. 
To do this, MC2 evaluates the expected reward of using the S2 solver in the current state to solve the given task, using information contained
in the model of self about past actions taken by this or other solvers to solve the same task, and the expected cost of running this solver.
MC2 then compares the expected reward for the S2 solver with the expected reward of the action proposed by the S1 solver: if the expected additional reward of running the S2 solver, as compared to using the S1 solution, is large enough, then MC2 activates the S2 solver. Otherwise, it adopts the S1 solution. 

To evaluate the expected reward of the action proposed by S1, MC2 retrieves from the model of self the expected immediate and future reward for the action in the current state 
(approximating the forward analysis to avoid a too costly computation),
and combines this information with the confidence the S1 solver has in the action.
The rationale for the behavior of MC2 is based on the design decision to avoid costly reasoning processes unless the additional cost is compensated by an even greater additional expected reward for the solution that the S2 solver will identify for this task. This is analogous to what happens in humans \cite{Shenhav2013}.



\section{Instantiating SOFAI on Grid Navigation}
\label{sec:sofai_def}





In the SOFAI instance that we consider and evaluate in this paper, the decision environment is a $9 \times 9$ grid and the task is to generate a trajectory from an initial state $S_0$ to a goal state $S_G$, by making moves from one state to an adjacent one in a sequence, while minimizing the penalties incurred. 

Such penalties are generated by constraints over moves (there are 8 moves for each state), specific states (grid cells), and state features (in our setting, these are colors associated to states). 
For example, there could be a penalty for moving left, for going to the cell (1,3), and for moving to a blue state.
In our specific experimental setting, 
any move brings a penalty of $-4$, each constraint violation gives a penalty of $-50$, and reaching the goal state gives a reward of $10$. 

\begin{figure}[t]
  \centering
  \includegraphics[height=4cm]{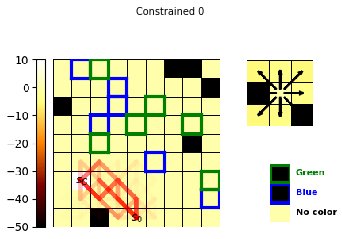}
\caption{Example of the constrained grid decision scenario. Black squares represents states with penalties. Penalties are generated also when the agent moves left or bottom-right, or when it moves to a blue or a green state. The red lines describe a set of trajectories generated by the agent (all with the same start and end point). The strength of the red color for each move corresponds to the amount of trajectories employing such move. } 
\label{fig:scobee}
\end{figure}

This decision environment is non-deterministic: 
there is a $10\%$ chance of failure, meaning that the decision of moving to a certain adjacent state may result in a move to another adjacent state chosen at random.
Figure \ref{fig:scobee} shows an example of our grid decision environment. 

Given this decision environment, we instantiate the SOFAI architecture as follows:
\begin{itemize}
\item one S1 solver, that uses information about the past trajectories to decide the next move (see below for details);
    \item one S2 solver, that uses MDFT to make the decision about the next move;
    \item MC agent: its behavior is described by Algorithm \ref{mc:pseudocode};
    \item model of the world: the grid environment;
    \item model of self: it includes past trajectories and their features (moves, reward, length, time);
    \item no model of others.
\end{itemize}




\begin{algorithm}
	\caption{The MC agent} 
	Input (Action a, Confidence c, State $s_x$, Partial Trajectory T)
	\label{mc:pseudocode}
	\begin{algorithmic}[1]

	    \IF
	    {$nTraj(s_x,ALL) \leq t_1$ \OR $\frac{partReward(T)}{avgReward(s_x)} \leq t_2$ \OR \\
	    $c \leq t_3$} 
	        \IF {$nTraj(s_x,S2) \leq t6$}
	            \STATE randomly adopt S1 decision or activate S2 solver
	        \ELSE
	            \STATE $expCost_{S2} \gets \frac{expTime_{S2}}{remTime}$
	            \IF {$expCost_{S2} \leq 1$ \AND \\ 
	            $\frac{(expReward_{S2}(s_x) – expReward(s_x,a))}{expCost_{S2}} > t_4$}
	                \STATE Set the attention weights in W
	                \STATE Activate the S2 solver
	            \ELSE
	                \STATE Adopt S1 decision
	            \ENDIF
	        \ENDIF
	    \ELSE
	        \STATE Adopt S1 decision
	    \ENDIF
	\end{algorithmic} 
\end{algorithm}

In Algorithm \ref{mc:pseudocode}:
\begin{itemize}
    \item 
$nTraj(s_x, \{S, ALL\})$ returns the number of times in state $s_x$ an action computed by solver $S$ ($ALL$ means any solver) has been adopted by the system;
if they are below $t_1$ (a natural number), it means that we don't have enough experience yet.
\item $partReward(T)$ and $avgReward(s_x)$ are respectively the  partial reward of the trajectory $T$ and the average partial reward that the agent gets when it usually reaches state $s_x$: if we are below $t_2$ (between $0$ and $1$), it means that we are performing worse than past experience.
\item $c$ is the confidence of the S1 solver: if it is below $t_3$ (between $0$ and $1$) it means that our attitude to risk does not tolerate the confidence level.
\end{itemize}

If any of the tests in line 1 are passed (meaning, the condition is not satisfied), the MC system (MC1) adopts the S1 decision.
Otherwise, it performs a more careful evaluation (MC2):
\begin{itemize}
\item 
$t_6$ checks that the S2 solver has enough experience. If not, a random choice between S1 and S2 is made (line 3).
\item Otherwise, 
it checks if it is convenient to activate S2 (line 6), comparing the expected gain in reward normalize by its cost. $t_4$ gives the tolerance for this gain. If it is convenient, MC activates the S2 solver (line 8), otherwise it adopts S1's decision. In this evaluation, 
$expTime_{S2}$ and $remTime$ are respectively the average amount of time taken by S2 to compute an action and the remaining time to complete the trajectory; $expReward_{S2}(s_x)$ and $expReward(s_x,a)$ are the expected reward using S2 in state $s_x$ and the expected reward of adopting action $a$ (computed by S1) in state $s_x$.
The expected reward for an action $a$ in a state $s_x$ is:
\[
 E(R|s_x, a) = \sum_{r_i \in R_{s_x,a}} P(r_i | s_x, a) * r_i
\]

where $R_{s_x,a}$ is the set of all the rewards in state $s_x$ taking the action $a$ that are stored in the model of self; $P(r_i | s_x, a)$ is the probability of getting the reward $r_i$ in state $s_x$ taking the action $a$. As the expected reward depends on the past experience stored in the model of self, it is possible to compute a confidence as follows:
\[
c(s_x, a) = sigmoid(\frac{(r - 0.5)}{ (\sigma + 1e-10))})
\]

where $\sigma$ is the standard deviation of the rewards in $s_x$ taking an action $a$, $r$ is the probability of taking action $a$ in state $s_x$.
\end{itemize}

MC1 and MC2 bear some resemblance to UCB and model-based learning in RL \cite{sutton-barto-rl}. However, in SOFAI we decompose some of these techniques to make decisions in a more fine grained manner.

The S1 agent, given a state $s_x$, chooses the action that maximizes the expected reward based on the past experience. That is:
$
\argmax_{a} (E(R|s_x, a)* c(s_x, a))
$.

The S2 agent, instead, employs the MDFT machinery (see Section \ref{sec:mdft-back}) to make a decision, where the $M$ matrix has two columns, containing the Q values of a nominal and constrained RL agents, and the attention weights $W$ are set in three possible ways: 1) attention to satisfying the constraints only if we have already violated many of them (denoted by 01), 2) attention to reaching the goal state only if the current partial trajectory is too long (denoted by 10), and 3) attention to both goal and constraints (denoted by 02). 
We will call the three resulting versions SOFAI 01, 10, and 02.



\section{Experimental Results}
\label{sec:experiments}


We generated at random $10$ grids, and for each grid we randomly chose: the initial and final states, 2 constrained actions, 6 constrained states, 12 constrained state features (6 green and  6 blue). For each grid, we run the following agents:
\begin{itemize}
    \item two reinforcement learning agents: one that tries to avoid the constraint penalties to reach the goal (called RL Constrained), and the other that just tries to reach the goal with no attention to the constraints (called RL Nominal). These agents will provide the baselines;
    \item the S1 solver;
    \item the S2 solver (that is, MDFT): this agent will be both a component of SOFAI and the provider of human-like trajectories;
    \item SOFAI 01, SOFAI 10, and SOFAI 02.
\end{itemize}
Each agent generates $1000$ trajectories.
We experimented with many combinations of values for the parameters. Here, we report the results for the following configuration: $t_1=200, t_2=0.8, t_3=0.4, t_4=0, t_6=1$.

\begin{figure}[h!]
  \centering
  \includegraphics[height=3cm]{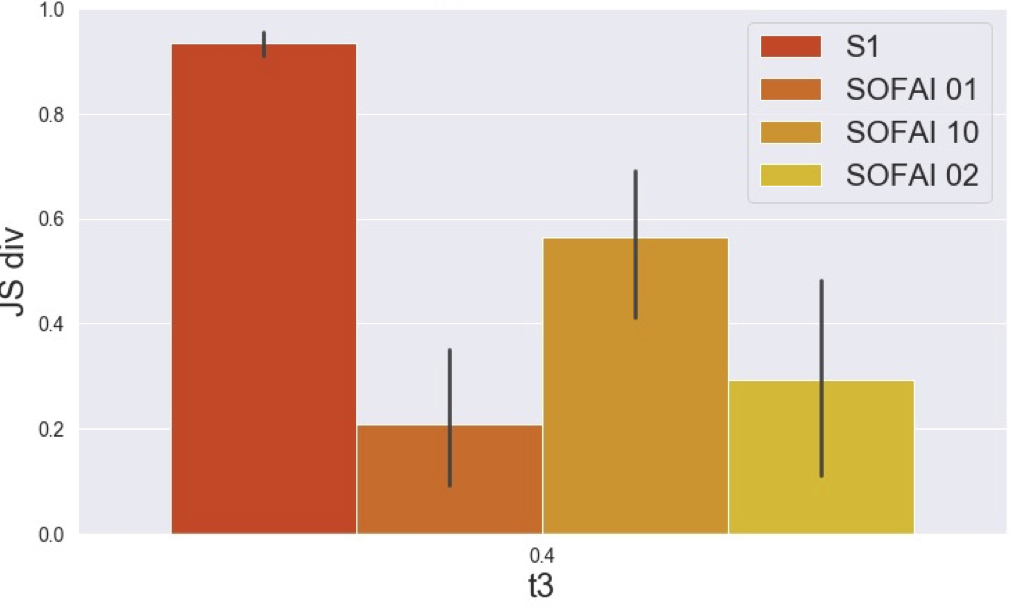}
\caption{Average JS divergence between the set of trajectories generated by MDFT and the other systems.} 
\label{fig:js_div}
\end{figure}

We first checked which agent generates trajectories that are more similar to the human ones (exemplified by MDFT).
Figure \ref{fig:js_div} reports the average JS-divergence between the set of trajectories generated by MDFT and the other systems. It is easy to see that SOFAI agents perform much better than S1, especially in the 01 configuration. 

\begin{figure*}[h!]
\begin{minipage}[t]{.33\textwidth}
  \centering
  \includegraphics[width=\linewidth]{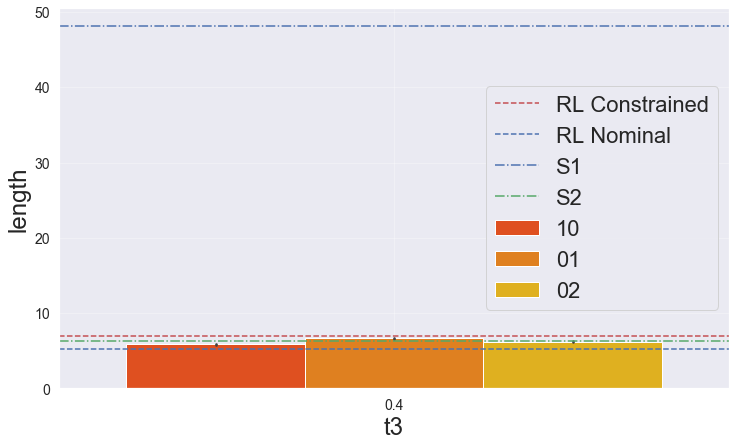}
\end{minipage}%
\hfill
\begin{minipage}[t]{.33\textwidth}
  \centering
  \includegraphics[width=\linewidth]{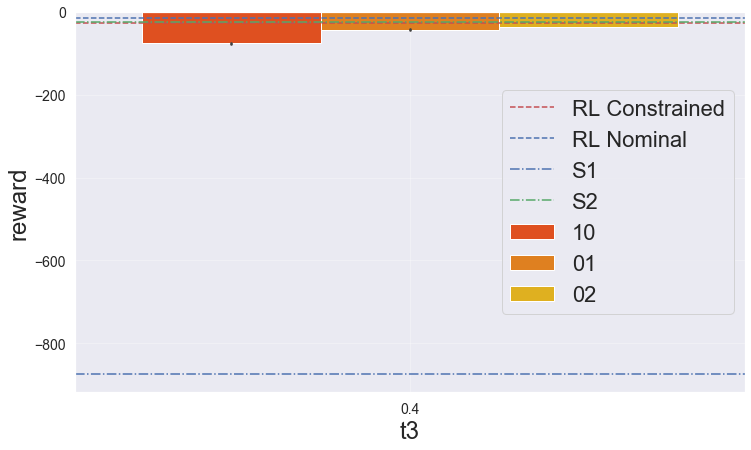}
\end{minipage}%
\hfill
\begin{minipage}[t]{.33\textwidth}
  \centering
  \includegraphics[width=\linewidth]{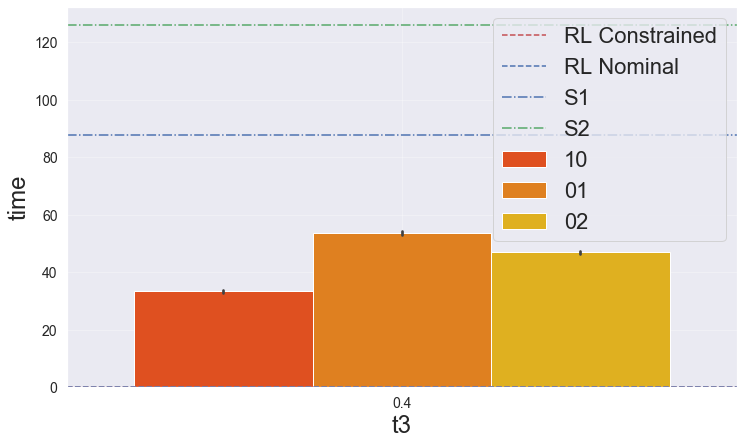}
\end{minipage}
\caption{Average length (left), reward (center), and time (right) for each trajectory, aggregated over 10 grids and 1000 trajectories.}
\label{fig:avg_values}
\end{figure*}

We then compared the three versions of SOFAI to S1 alone, S2 alone, and the two RL agents, in terms of  the length of the generated paths, total reward, and time to generate the trajectories, see Figure \ref{fig:avg_values}.
It is easy to see that S1 performs very badly on all three criteria, while the other systems are comparable. Notice that RL Nominal represents a lower bound for the length criteria and an upper bound for the reward, since it gets to the goal with no attention to satisfying the constraints.
For both reward and time, SOFAI (which combines S1 and S2) performs better than using only S1 or only S2. 

\begin{figure*}[h!]
\begin{minipage}[t]{.33\textwidth}
  \centering
  \includegraphics[width=\linewidth]{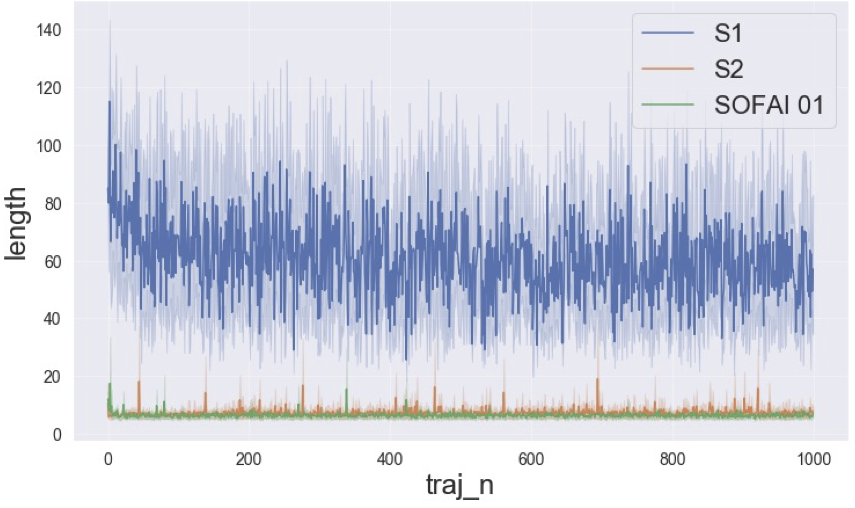}
\end{minipage}%
\hfill
\begin{minipage}[t]{.33\textwidth}
  \centering
  \includegraphics[width=\linewidth]{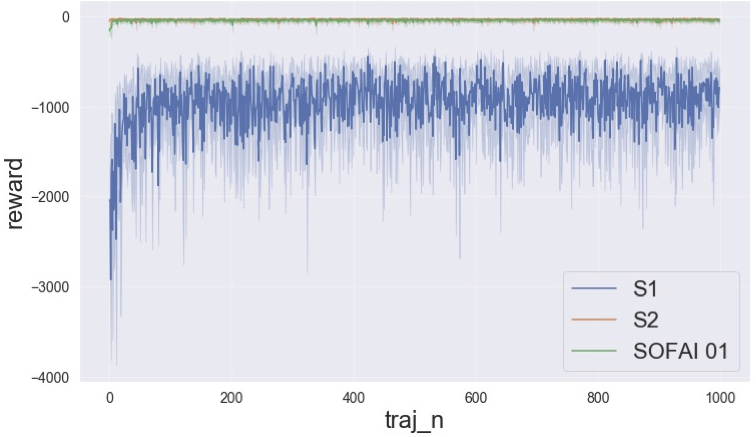}
\end{minipage}%
\hfill
\begin{minipage}[t]{.33\textwidth}
  \centering
  \includegraphics[width=\linewidth]{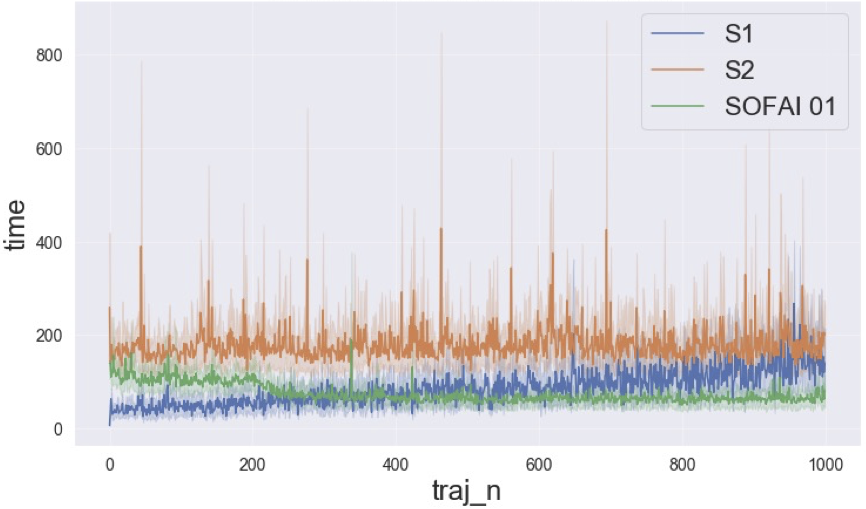}
\end{minipage}
\caption{Average length (left), reward (center), and time to compute each trajectory (right),  aggregated over 10 grids.}
\label{fig:time_analysis_per_system}
\end{figure*}

\begin{figure*}[h!]
\begin{minipage}[t]{.33\textwidth}
  \centering
  \includegraphics[width=\linewidth]{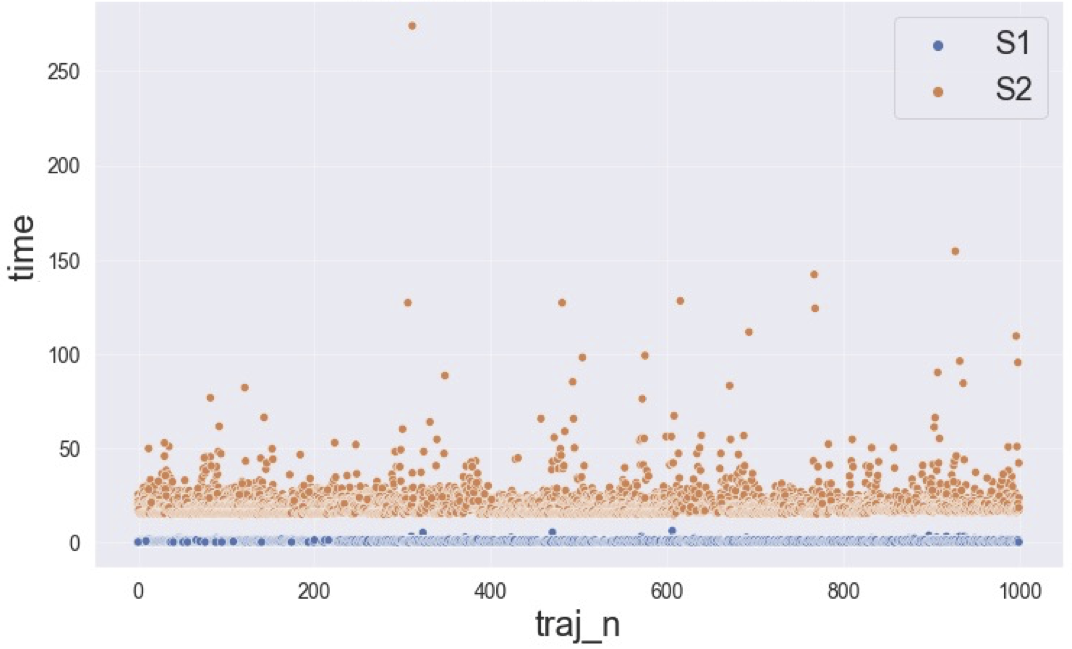}
\end{minipage}%
\hfill
\begin{minipage}[t]{.33\textwidth}
  \centering
  \includegraphics[width=\linewidth]{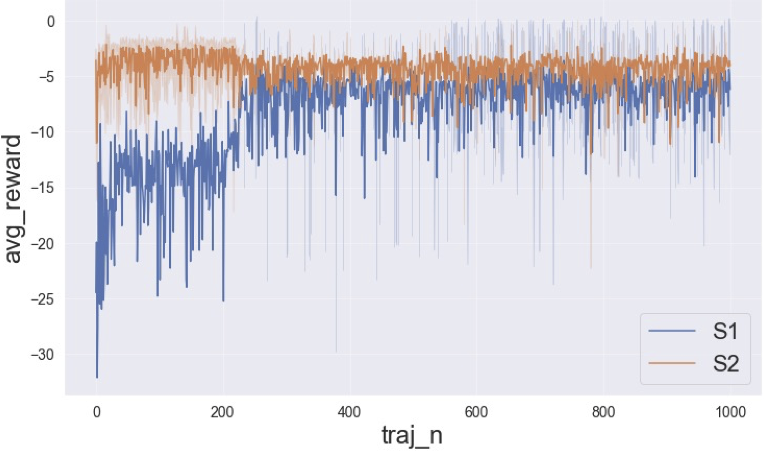}
\end{minipage}%
\hfill
\begin{minipage}[t]{.33\textwidth}
  \centering
  \includegraphics[width=\linewidth]{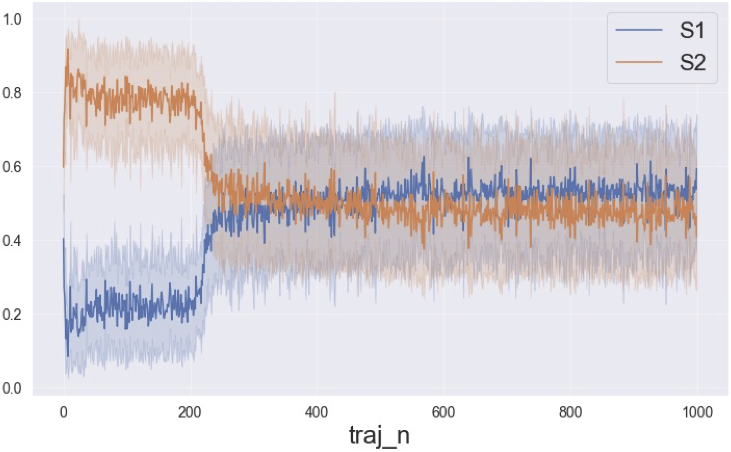}
\end{minipage}
\caption{Time to compute a move (left), average reward for a move (center), and average fraction of times each sub-system is used (right), over 10 grids.}
\label{fig:time_analysis}
\end{figure*}



We then passed from the aggregate results over all $1000$ trajectories to checking the behavior of SOFAI and the other agents over time, from trajectory 1 to 1000.
The goal is to see how SOFAI methods evolve in their behavior and their decisions on how to combine its S1 and S2 agents.
Given that SOFAI 01 performs comparably or better than the other two versions, in the following experimental results we will only show the behavior of this version and will denote it simply as SOFAI.


Figure \ref{fig:time_analysis_per_system} shows the length, reward, and time for each of the 1000 trajectories, comparing SOFAI to S1 and to S2. 
In terms of  length and reward, S1 does not perform well at all, while SOFAI and S2 are comparable. However, the time chart shows that SOFAI is much faster than S2 and over time it also becomes faster than S1, even if it uses a combination of S1 and S2. This is due to the fact that S1 alone cannot exploit the experience gathered by S2 within SOFAI, so it generates much worse and longer trajectories, which require much more time.
Perhaps the most interesting is
Figure \ref{fig:time_analysis}. The left figure shows the average time spent by S1 and S2 within SOFAI in taking a single decision (thus a single move in the trajectory): S2 always takes more time than S1, and this is stable over time.
The center figure shows the average reward for a single move: S2 is rather stable in generating high quality moves, while S1 at first performs very badly (since there is not enough experience yet) and later generates better moves (but still worse than S2). The question is now: how come S1 improves so much over time? The answer is given by the right figure which shows the percentage of usage of S1 and S2 in each trajectory. As we can see, at the beginning SOFAI uses mostly S2, since the lack of experience makes S1 not trustable (that is, the MC algorithm does not lead to the adoption of the S1 decision). After a while, with enough trajectories built by (mostly) S2 and stored in the model of self, SOFAI (more precisely, the MC agent) can trust S1 enough to use it more often when deciding the next move, so much that after about 450 trajectories S1 is used more often than S2. This allows SOFAI to be faster while not degrading the reward of the generated trajectories. This behavior is similar to what happens in humans (as described in Section \ref{fshumans}): we first tackle a non-familiar problem with our System 2, until we have enough experience that it becomes familiar and we pass to using System 1.

\section{Future Work}

We presented SOFAI, a conceptual architecture inspired by the thinking fast and slow theory of human decision making, and we described its behavior over a grid environment, showing that it is able to combine S1 and S2 decision modalities to generate high quality decisions faster than using just S1 or S2.
We plan to generalize our work to allow for several S1 and/or S2 solvers and several problems for the same architecture, thus tackling issues of ontology and similarity. 

\section{Acknowledgements}
We would like to thank Daniel Kahneman for his continuous support for our work, and many enlightening discussions. We also would like to thank Aanya Khandelwal (Georgia Tech) for her contributions to the design of the metacognition module for the grid environment during her 2021 internship at IBM, as well as all the other project team members (Grady Booch, Kiran Kate, Nick Linck, Keerthiram Murugesan, Mattia Rigotti, at IBM) for extensive discussions on both the theoretical and the experimental part of this work.

\newpage
\bibliographystyle{named}
\bibliography{biblio}

\begin{thebibliography}{}

\bibitem[\protect\citeauthoryear{Ackerman and Thompson}{2017}]{Ackerman2017}
Rakefet Ackerman and Valerie~A Thompson.
\newblock Meta-reasoning: Monitoring and control of thinking and reasoning.
\newblock {\em Trends in Cognitive Sciences}, 21(8):607--617, 2017.

\bibitem[\protect\citeauthoryear{Anthony \bgroup \em et al.\egroup
  }{2017}]{anthony2017thinking}
Thomas Anthony, Zheng Tian, and David Barber.
\newblock Thinking fast and slow with deep learning and tree search.
\newblock In {\em Advances in Neural Information Processing Systems}, pages
  5360--5370, 2017.

\bibitem[\protect\citeauthoryear{Bengio}{2017}]{bengio2017consciousness}
Yoshua Bengio.
\newblock The consciousness prior.
\newblock {\em arXiv preprint arXiv:1709.08568}, 2017.

\bibitem[\protect\citeauthoryear{Booch \bgroup \em et al.\egroup
  }{2021}]{aaai2021-blue}
Grady Booch, Francesco Fabiano, Lior Horesh, Kiran Kate, Jonathan Lenchner,
  Nick Linck, Andreas Loreggia, Keerthiram Murgesan, Nicholas Mattei, Francesca
  Rossi, and Biplav Srivastava.
\newblock Thinking fast and slow in {AI}.
\newblock In {\em Proceedings of the AAAI Conference on Artificial
  Intelligence}, volume~35, pages 15042--15046, 2021.

\bibitem[\protect\citeauthoryear{Busemeyer and
  Townsend}{1993}]{busemeyer1993decision}
Jerome~R Busemeyer and James~T Townsend.
\newblock Decision field theory: a dynamic-cognitive approach to decision
  making in an uncertain environment.
\newblock {\em Psychological review}, 100(3):432, 1993.

\bibitem[\protect\citeauthoryear{Carruthers}{2021}]{Carruthers}
Peter Carruthers.
\newblock Explicit nonconceptual metacognition.
\newblock {\em Philosophical Studies}, 178(7):2337--2356, 2021.

\bibitem[\protect\citeauthoryear{Chen \bgroup \em et al.\egroup
  }{2019}]{chen2019deep}
Di~Chen, Yiwei Bai, Wenting Zhao, Sebastian Ament, John~M Gregoire, and Carla~P
  Gomes.
\newblock Deep reasoning networks: Thinking fast and slow.
\newblock {\em arXiv preprint arXiv:1906.00855}, 2019.

\bibitem[\protect\citeauthoryear{Cox and Raja}{2011}]{CoxRaja2011}
Michael~T Cox and Anita Raja.
\newblock {\em Metareasoning: Thinking about thinking}.
\newblock MIT Press, 2011.

\bibitem[\protect\citeauthoryear{Cox}{2005}]{Cox2005}
Michael~T Cox.
\newblock Metacognition in computation: A selected research review.
\newblock {\em Artificial intelligence}, 169(2):104--141, 2005.

\bibitem[\protect\citeauthoryear{Flavell}{1979}]{Flavell1979}
John~H Flavell.
\newblock Metacognition and cognitive monitoring: A new area of
  cognitive--developmental inquiry.
\newblock {\em American psychologist}, 34(10):906, 1979.

\bibitem[\protect\citeauthoryear{Gigerenzer and
  Brighton}{2009}]{gigerenzer2009}
Gerd Gigerenzer and Henry Brighton.
\newblock Homo heuristicus: Why biased minds make better inferences.
\newblock {\em Topics in Cognitive Science}, 1(1):107--143, 2009.

\bibitem[\protect\citeauthoryear{Goel \bgroup \em et al.\egroup
  }{2017}]{goel2017thinking}
Gautam Goel, Niangjun Chen, and Adam Wierman.
\newblock Thinking fast and slow: Optimization decomposition across timescales.
\newblock In {\em IEEE 56th Conference on Decision and Control (CDC)}, pages
  1291--1298. IEEE, 2017.

\bibitem[\protect\citeauthoryear{Gulati \bgroup \em et al.\egroup
  }{2020}]{gulati2020interleaving}
Aditya Gulati, Sarthak Soni, and Shrisha Rao.
\newblock Interleaving fast and slow decision making.
\newblock {\em arXiv preprint arXiv:2010.16244}, 2020.

\bibitem[\protect\citeauthoryear{Hotaling \bgroup \em et al.\egroup
  }{2010}]{hotaling2010theoretical}
Jared~M Hotaling, Jerome~R Busemeyer, and Jiyun Li.
\newblock Theoretical developments in decision field theory: Comment on
  tsetsos, usher, and chater (2010).
\newblock {\em Psychological Review}, 2010.

\bibitem[\protect\citeauthoryear{Kahneman}{2011}]{kahneman2011thinking}
Daniel Kahneman.
\newblock {\em Thinking, Fast and Slow}.
\newblock Macmillan, 2011.

\bibitem[\protect\citeauthoryear{Kerschke \bgroup \em et al.\egroup
  }{2019}]{kerschke2019automated}
Pascal Kerschke, Holger~H Hoos, Frank Neumann, and Heike Trautmann.
\newblock Automated algorithm selection: Survey and perspectives.
\newblock {\em Evolutionary computation}, 27(1):3--45, 2019.

\bibitem[\protect\citeauthoryear{Kim \bgroup \em et al.\egroup
  }{2019}]{Kim2019}
Dongjae Kim, Geon~Yeong Park, PO~John, Sang~Wan Lee, et~al.
\newblock Task complexity interacts with state-space uncertainty in the
  arbitration between model-based and model-free learning.
\newblock {\em Nature communications}, 10(1):1--14, 2019.

\bibitem[\protect\citeauthoryear{Kralik and et al.}{2018}]{Kralik2018}
Jerald~D Kralik and et~al.
\newblock Metacognition for a common model of cognition.
\newblock {\em Procedia computer science}, 145:730--739, 2018.

\bibitem[\protect\citeauthoryear{Littman and et al.}{2021}]{ai1002021}
Michael~L. Littman and et~al.
\newblock {Gathering Strength, Gathering Storms: The One Hundred Year Study on
  Artificial Intelligence (AI100) 2021 Study Panel Report}.
\newblock {\em Stanford University}, 2021.

\bibitem[\protect\citeauthoryear{Marcus}{2020}]{marcus2020next}
Gary Marcus.
\newblock The next decade in {AI}: {F}our steps towards robust artificial
  intelligence.
\newblock {\em arXiv preprint arXiv:2002.06177}, 2020.

\bibitem[\protect\citeauthoryear{Mittal \bgroup \em et al.\egroup
  }{2017}]{mittal2017thinking}
Sudip Mittal, Anupam Joshi, and Tim Finin.
\newblock Thinking, fast and slow: Combining vector spaces and knowledge
  graphs.
\newblock {\em arXiv preprint arXiv:1708.03310}, 2017.

\bibitem[\protect\citeauthoryear{Nelson}{1990}]{NN1990}
Thomas~O Nelson.
\newblock Metamemory: A theoretical framework and new findings.
\newblock In {\em Psychology of learning and motivation}, volume~26, pages
  125--173. Elsevier, 1990.

\bibitem[\protect\citeauthoryear{Noothigattu and et
  al.}{2019}]{DBLP:journals/ibmrd/NoothigattuBMCM19}
R.~Noothigattu and et~al.
\newblock Teaching {AI} agents ethical values using reinforcement learning and
  policy orchestration.
\newblock {\em {IBM} J. Res. Dev.}, 63(4/5):2:1--2:9, 2019.

\bibitem[\protect\citeauthoryear{Posner}{2020}]{Posner2020}
Ingmar Posner.
\newblock Robots thinking fast and slow: On dual process theory and
  metacognition in embodied {AI}.
\newblock 2020.

\bibitem[\protect\citeauthoryear{Proust}{2013}]{Proust2004}
Jo{\"e}lle Proust.
\newblock {\em The philosophy of metacognition: Mental agency and
  self-awareness}.
\newblock OUP Oxford, 2013.

\bibitem[\protect\citeauthoryear{Roe \bgroup \em et al.\egroup
  }{2001}]{roe2001multialternative}
Robert~M Roe, Jermone~R Busemeyer, and James~T Townsend.
\newblock Multialternative decision field theory: A dynamic connectionst model
  of decision making.
\newblock {\em Psychological review}, 108(2):370, 2001.

\bibitem[\protect\citeauthoryear{Rossi and Mattei}{2019}]{RoMa19a}
F.~Rossi and N.~Mattei.
\newblock Building ethically bounded {AI}.
\newblock In {\em Proceedings of the 33rd AAAI Conference on Artificial
  Intelligence (AAAI)}, 2019.

\bibitem[\protect\citeauthoryear{Shenhav \bgroup \em et al.\egroup
  }{2013}]{Shenhav2013}
Amitai Shenhav, Matthew~M Botvinick, and Jonathan~D Cohen.
\newblock The expected value of control: an integrative theory of anterior
  cingulate cortex function.
\newblock {\em Neuron}, 79(2):217--240, 2013.

\bibitem[\protect\citeauthoryear{Sutton and Barto}{2018}]{sutton-barto-rl}
Richard~S. Sutton and Andrew~G. Barto.
\newblock {\em Reinforcement Learning: An Introduction, 2nd Edition}.
\newblock A Bradford Book, Cambridge, MA, USA, 2018.

\bibitem[\protect\citeauthoryear{Thompson \bgroup \em et al.\egroup
  }{2011}]{Thompson2011}
Valerie~A Thompson, Jamie A~Prowse Turner, and Gordon Pennycook.
\newblock Intuition, reason, and metacognition.
\newblock {\em Cognitive psychology}, 63(3):107--140, 2011.

\end{thebibliography}

\end{document}